\documentclass[10pt, a4paper]{article}
\usepackage{lrec}
\usepackage{graphicx}
\usepackage{tabularx}
\usepackage{soul}
\usepackage{booktabs}

\usepackage{epstopdf}
\usepackage[latin1]{inputenc}

\usepackage{hyperref}
\usepackage{xstring}

\newcommand\blfootnote[1]{%
  \begingroup
  \renewcommand\thefootnote{}\footnote{#1}%
  \addtocounter{footnote}{-1}%
  \endgroup
}

\makeatletter
\renewcommand{\p@section}{\arabic{section}\expandafter\@gobble}
\renewcommand{\p@subsection}{\thesection\arabic{subsection}\expandafter\@gobble}
\renewcommand{\p@subsubsection}{\thesubsection\arabic{subsubsection}\expandafter\@gobble}
\makeatother

\title{A Resource for Computational Experiments on Mapudungun}

\name{Mingjun Duan\textsuperscript{$\dagger$1}, Carlos Fasola\textsuperscript{$\dagger$2}, Sai Krishna Rallabandi\textsuperscript{1}, Rodolfo M. Vega\textsuperscript{1},\\\bf\large  Antonios Anastasopoulos\textsuperscript{1}, Lori Levin\textsuperscript{1}, Alan W Black\textsuperscript{1}}

\address{\begin{tabular}{cc}
    \textsuperscript{1}Language Technologies Institute &  \textsuperscript{2}Interpreting and Translation Studies\\
    Carnegie Mellon University & Wake Forest University\\
    5000 Forbes Ave, Pittsburgh PA 15213, USA & 1834 Wake Forest Road, Winston-Salem, NC 27109, USA\\
\end{tabular}\\
         \{mingjund,srallaba,rmvega\}@andrew.cmu.edu, fasolaca@wfu.edu,\\
         \{aanastas, levin, awb\}@cs.cmu.edu\\
         }

\abstract{
We present a resource for computational experiments on Mapudungun, a polysynthetic indigenous language spoken in Chile with upwards of 200 thousand speakers.  We provide~142 hours of culturally significant conversations in the domain of medical treatment.  The conversations are fully transcribed and translated into Spanish.  The transcriptions also include annotations for code-switching and non-standard pronunciations. We also provide baseline results on three core NLP tasks: speech recognition, speech synthesis, and machine translation between Spanish and Mapudungun. We further explore other applications for which the corpus will be suitable, including the study of code-switching, historical orthography change, linguistic structure, and sociological and anthropological studies.\\ 
\newline \Keywords{Mapudungun, Endangered Languages, Speech, Machine Translation} }

\begin{document}

\maketitleabstract

\section{Introduction}
\blfootnote{$^\dagger$Equal contribution}Recent years have seen unprecedented progress for Natural Language Processing (NLP) on almost every NLP subtask. Even though low-resource settings have also been explored, this progress has overwhelmingly been observed in languages with significant data resources that can be leveraged to train deep neural networks.  Low-resource languages still lag behind.  

Endangered languages pose an additional challenge.  The process of documenting an endangered language typically includes the creation of word lists, audio and video recordings, notes, or grammar fragments, with the created resources then stored in large online linguistics archives.  This process is often hindered by the \textit{Transcription Bottleneck}: the linguistic fieldworker and the language community may not have time to transcribe all of the recordings and may only transcribe segments that are linguistically salient for publication or culturally significant for the creation of community resources.   

With this work we make publicly available a large corpus in Mapudungun, a language of the indigenous Mapuche people of southern Chile and western Argentina.  We hope to ameliorate the resource gap and the transcription bottleneck in two ways. First, we are providing a larger data set than has previously been available, and second, we are providing baselines for NLP tasks (speech recognition, speech synthesis, and machine translation).   In providing baselines and datasets splits, we hope to further facilitate research on low-resource NLP for this language through our data set.   Research on low-resource speech recognition is particularly important in relieving the transcription bottleneck, while tackling the research challenges that speech synthesis and machine translation pose for such languages could lead to such systems being deployed to serve more under-represented communities.  

\section{The Mapudungun Language}
Mapudungun (iso 639-3: arn) 
is an indigenous language of the Americas spoken natively in Chile and Argentina, with an estimated 100 to 200 thousand speakers in Chile and 27 to 60 thousand speakers in Argentina \cite[41--3]{zuniga:2006}. 
It is an isolate language and is classified as threatened by Ethnologue, hence the critical importance of all documentary efforts.
Although the morphology of nouns is relatively simple, Mapudungun verb morphology is highly agglutinative and complex. Some analyses provide as many as~36 verb suffix slots~\cite{smeets1989mapuche}. A typical complex verb form occurring in our corpus of spoken Mapudungun consists of five or six morphemes. 


Mapudungun has several interesting grammatical properties. It is a polysynthetic language in the sense of \cite{baker:1996}; see \cite{loncon:2011} for explicit argumentation. 
As with other polysynthetic languages, Mapudungun has Noun Incorporation; however, it is unique insofar as the Noun appears to the right of the Verb, instead of to the left, as in most polysynthetic languages \cite{bakeretal:2005}.
One further distinction of Mapudungun is that, whereas other polysynthetic languages are characterized by a lack of infinitives, Mapudungun has infinitival verb forms; that is, while subordinate clauses in Mapudungun closely resemble possessed nominals and may occur with an analytic marker resembling possessor agreement, there is no agreement inflection on the verb itself.
One further remarkable property of Mapudungun is its inverse voice system of agreement, whereby the highest agreement is with the argument highest in an animacy hierarchy regardless of thematic role \cite{arnold:1996}.

\section{Related Work}
Mapudungun grammar has been studied since the arrival of European missionaries and colonizers hundreds of years ago.   More recent descriptions of Mapudungun grammar~\cite{smeets1989mapuche} and~\cite{zuniga:2006} informed the collection of the resource that we are presenting in this paper.


Portions of our resource have been used in early efforts to build language systems for Mapudungun. In particular, \cite{monson-etal-2004-data} focused on Mapudungun morphology in order to create spelling correction systems, while \cite{Monson-etal-2006-saltmil}, \cite{Levin-et-al-2002}, \cite{Font-Liitjos-et-al-2005}, and \cite{monson-etal-2008-linguistic} developed hybrid rule- and phrase-based Statistical Machine Translation systems.

Naturally, similar works in collecting corpora in Indigenous languages of Latin America are abundant, but very few, if any, have the scale and potential of our resource to be useful in many downstream language-specific and inter-disciplinary applications. A general overview of the state of NLP for the under-represented languages of the Americas can be found at \cite{mager2018challenges}. To name a few of the many notable works, \cite{montanomixtec} created a parallel Mixtec-Spanish corpus for Machine Translation and \cite{kazeminejad-etal-2017-creating} created lexical resources for Arapaho, while \cite{cardenassiminchik} and \cite{cavar2016endangered} focused on building speech corpora for Southern Quechua and Chatino respectively.

\section{The Resource}
\begin{table}[]
    \centering
    \begin{tabular}{c|cccc}
    \toprule
         & Train & Dev & Test & All  \\
    \midrule
    \multicolumn{5}{l}{\textit{Audio}}\\
        Hours & 117 & 4 & 21 & 142\\
        Conversations & 285 & 12 & 46 & 343 \\
        Turns & 78.4k & 2.5k & 14k & 94.9k \\
    \multicolumn{5}{l}{\textit{Mapudungun Transcriptions}}\\
        Sentences & 221.8k & 8.2k & 36.3k & 266.3k\\
        Tokens (arn) & 781k & 31.2k & 126k & 938.3k \\
        Types (arn)& 92.2k & 7.1k & 22.3k & 121.6k \\
    \multicolumn{5}{l}{\textit{Spanish Translations}}\\
        Tokens (spa) & 1,029k & 38.9k & 164k & 1,232k\\
        Types (spa)& 24.6k & 3.7k & 8.6k & 37k\\
    \bottomrule
    \end{tabular}
    \caption{Basic Statistics of our corpus.}
    \label{tab:statistics}
\end{table}

The resource\footnote{Project page: \url{http://tts.speech.cs.cmu.edu/mapudungun/}. Data also available at \url{https://github.com/mingjund/mapudungun-corpus}} is comprised of~142 hours of spoken Mapudungun that was recorded during the AVENUE project~\cite{Levin-et-al-2002} in 2001 to 2005. The data was recorded under a partnership between the AVENUE project, funded by the US National Science Foundation at Carnegie Mellon University, the Chilean Ministry of Education (Mineduc), and the Instituto de Estudios Ind\'{i}genas at Universidad de La Frontera, originally spanning 170 hours of audio. We have recently cleaned the data and are releasing it publicly for the first time (although it has been shared with individual researchers in the past) along with NLP baselines.  


The recordings were transcribed and translated into Spanish at the Instituto de Estudios Ind\'{i}genas at Universidad de La Frontera. The corpus covers three dialects of Mapudungun: about 110 hours of Nguluche, 20 hours of Lafkenche and 10 hours of Pewenche.
The three dialects are quite similar, with some minor semantic and phonetic differences. The fourth traditionally distinguished dialect, Huilliche, has several grammatical differences from the other three and is classified by Ethnologue as a separate language, iso 639-3: huh, and as nearly extinct.

The recordings are restricted to a single domain: primary, preventive, and treatment health care, including both Western and Mapuche traditional medicine. The recording sessions were conducted as interactive conversations so as to be natural in Mapuche culture,  and they were open-ended, following an ethnographic approach. The interviewer was trained in these methods along with the use of the digital recording systems that were available at the time. We also followed human subject protocol. Each person signed a consent form to release the recordings for research purposes and the data have been accordingly anonymized. Because Machi (traditional Mapuche healers) 
were interviewed, we asked the transcribers to delete any culturally proprietary knowledge that a Machi may have revealed during the conversation. Similarly, we deleted any names or any information that may identify the participants.

The corpus is culturally relevant because it was created by Mapuche people, using traditional ways of relating to each other in conversations. They discussed personal experiences with primary health care in the traditional Mapuche system and the Chilean health care system, talking about illnesses and the way they were cured. The participants ranged from~16 years old to~100 years old, almost in equal numbers of men and women, and they were all native speakers of Mapudungun.  



\paragraph{Orthography}
At the time of the collection and transcription of the corpus, the orthography of Mapudungun was not standardized.
The Mapuche team at the Instituto de Estudios Ind\'{i}genas (IEI -- Institute for Indigenous Studies) developed a supra-dialectal alphabet that comprises 28 letters that cover 32 phones used in the three Mapudungun variants. The main criterion for choosing alphabetic characters was to use the current Spanish keyboard that was available on all computers in Chilean offices and schools. The alphabet used the same letters used in Spanish for those phonemes that sound like Spanish phonemes. Diacritics such as apostrophes were used for sounds that are not found in Spanish.

As a result, certain orthographic conventions that were made at the time deviate from the now-standard orthography of Mapudungun, Azumchefe.
We plan to normalize the orthography of the corpus, and in fact a small sample has already been converted to the modern orthography.
However, we believe that the original transcriptions will also be invaluable for academic, historical, and cultural purposes, hence we release the corpus using these conventions. 
\begin{table*}[t]
    \centering
    \begin{tabular}{lp{13cm}}
    \toprule
        \multicolumn{2}{l}{Conversation: nfmcp-nmedp2} \\
        \multicolumn{2}{l}{Speaker (Start -- End)} \\
    \midrule
        \multicolumn{2}{l}{nfmcp: 9min 27.069sec -- 10min 6.102sec} \\
        \multicolumn{1}{r}{Transcription:} & kuydaw\"{u}nmu fey tuntenmurume fey m\"{u}le rume kutran fey kuydawenulmu weche, wechengele che, wechengele ki\~{n}e domo ki\~{n}e wentru ka famechi tui m\"{u}ten kutran fey kuydawununmu m\"{u}ten \texttt{[SPA]}pero kuydawetulele fey newe kontulay kutran \texttt{[SPA]}pues \texttt{[!1pu']} welu, kuydanunu m\"{u}tewe rangi\~{n} wentru, rangin domolele deuma ki\~{n}e che fey rume kuntueyu ta chi kutran kuydaw\"{u}nmu kuydaw\"{u}nmu fey newe kontulayu kutran \\
        \multicolumn{1}{r}{Translation:} & Por no cuidarse hay en cualquier tiempo aparece de repente la enfermedad eso es porque la gente j\'{o}ven no se cuida si es joven la persona si es joven una mujer un hombre as\'{i} tambi\'{e}n toma no m\'{a}s la enfermedad eso es por no cuidarse no m\'{a}s pero si se vuelve a cuidar ah\'{i} no le entra mucho la enfermedad pero si no se cuida mucho cuando la persona este en la edad mediana hombre o mujer la da mucho esta enfermedad por no cuidarse, por no cuidarse ah\'{i} no le entra mucho esa enfermedad.\\
    \bottomrule
    \end{tabular}
    \caption{Example of an utterance along with the different annotations. We additionally \texttt{highlight} the code-switching annotations (\texttt{[SPA]} indicates Spanish words) as well as pre-normalized transcriptions that indicating non-standard pronunciations (\texttt{[!1pu']} indicates that the previous~1 word was pronounced as `\textit{pu'}' instead of `\textit{pues}').}
    \label{tab:example}
\end{table*}

\paragraph{Additional Annotations}
In addition, the transcription includes annotations for noises and disfluencies including aborted words, mispronunciations, poor intelligibility, repeated and corrected words, false starts, hesitations, undefined sound or pronunciations, non-verbal articulations, and pauses. Foreign words, in this case Spanish words, are also labelled as such.

\paragraph{Cleaning}
The dialogues were originally recorded using a Sony DAT recorder (48kHz), model TCD-D8, and Sony digital stereo microphone, model ECM-DS70P.
Transcription was performed with the TransEdit transcription tool v.1.1 beta 10\footnote{developed by Susanne Burger and Uwe Meier}, which synchronizes the transcribed text and the wave files.

However, we found that a non-trivial number of the utterance boundaries and speaker annotations were flawed.  Also some recording sessions did not have a complete set of matching audio, transcription, and translation files.
Hence, in an effort to provide a relatively ``clean" corpus for modern computational experiments, we converted the encoding of the textual transcription from Latin-1 to Unicode, DOS to UNIX line endings, a now more standard text encoding format than what was used when the data was first collected.
Additionally, we renamed a small portion of files which had been misnamed and removed several duplicate files.

Although all of the data was recorded with similar equipment in relatively quiet environments, the acoustics are not as uniform as we would like for building speech synthesizers.  Thus we applied standardized power normalization. We also moved the boundaries of the turns to standardize the amount of leading and trailing silence in each turn. This is a standard procedure for speech recognition and synthesis datasets.
Finally we used the techniques in \cite{black2019wilderness} for found data to re-align the text to the audio and find out which turns are best (or worst) aligned so that we can select segments that give the most accurate alignments.   Some of the misalignments may in part be due to varied orthography, and we intend, but have not yet, to investigate normalization of orthography (i.e. spelling correction) to mitigate this.

\paragraph{Training, Dev, and Test Splits}
We created two training sets, one appropriate for single-speaker speech synthesis experiments, and one appropriate for multiple-speaker speech recognition and machine translation experiments. In both cases, our training, development, and test splits are performed at the dialogue level, so that all examples from each dialogue belong to exactly one of these sets. 

For single-speaker speech synthesis, we only used the dialog turns of the speaker with the largest volume of data (\texttt{nmlch} -- one of the interviewers). The training set includes~$221.8$ thousand sentences from~285 dialogues, with~12 and~46 conversations reserved for the development and test set.

For speech recognition experiments, we ensured that our test set includes unique speakers as well as speakers that overlap with the training set, in order to allow for comparisons of the ability of the speech recognition system to generalize over seen and new speakers. For consistency, we used the same dataset splits for the machine translation experiments. The statistics in Table~\ref{tab:statistics} reflect this split.

\section{Applications}
Our resource has the potential to be the basis of computational research in Mapudungun across several areas.
Since the collected audio has been transcribed, our resource is appropriate for the study of automatic speech recognition and speech synthesis. The Spanish translations enable the creation of machine translation systems between Mapudungun and Spanish, as well as end-to-end (or direct) speech translation.
We in fact built such speech synthesis, speech recognition, and machine translation systems as a showcase of the usefulness of our corpus in that research direction.

Furthermore, our annotations of the Spanish words interspersed in Mapudungun speech could allow for a study of code-switching patterns within the Mapuche community. 7.3\% of the transcription tokens are in Spanish, and the remaining are in Mapudungun.
In addition, our annotations of non-standardized orthographic transcriptions could be extremely useful in the study of historical language and orthography change as a language moves from predominantly \textit{oral} to being written in a standardized orthography, as well as in building spelling normalization and correction systems.
The relatively large amount of data that we collected will also allow for the training of large language models, which in turn could be used as the basis for predictive keyboards tailored to Mapudungun.
Last, since all data are dialogues annotated for the different speaker turns, they could be useful for building Mapudungun dialogue systems and chatbot-like applications.

The potential applications of our resource, however, are not exhausted in language technologies.
The resource as a whole could be invaluable for ethnographic and sociological research, as the conversations contrast traditional and Western medicine practices, and they could reveal interesting aspects of the Mapuche culture.  

In addition, the corpus is a goldmine of data for studying the morphostyntax of Mapudungun~\cite{fasola2015topics}. As an isolate polysynthetic language, the study of Mapudungun can provide insights into the range of possibilities within human languages can work. 

\section{Baseline Results}

Using the aforementioned higher quality portions of the corpus, we trained baseline systems for Mapudungun speech recognition and speech synthesis, as well as Machine Translation systems between Mapudungun and Spanish.

\paragraph{Speech Synthesis}

In our previous work on building speech systems on found data in 700 languages, \cite{black2019wilderness}, we addressed alignment issues (when audio is not segmented into turn/sentence sized chunks) and correctness issues (when the audio does not match the transcription).  We used the same techniques here, as described above.    

For the best quality speech synthesis we need a few hours of phonetically-balanced, single-speaker, read speech.  
Our first step was to use the start and end points for each turn in the dialogues, and select those of the most frequent speaker, \texttt{nmlch}.
This gave us around 18250 segments.  We further automatically removed excessive silence from the start, middle and end of these turns (based on occurrence of F0). This gave us 13 hours and 48 minutes of speech.  

We phonetically aligned this data and built a speech clustergen statistical speech synthesizer \cite{clustergen06} from all of this data. We resynthesized all of the data and measured the difference between the synthesized data and the original data using Mel Cepstral Distortion, a standard method for automatically measuring quality of speech generation \cite{kominek08}.  We then ordered the segments by their generation score and took the top 2000 turns to build a new synthesizer, assuming the better scores corresponded to better alignments, following the techniques of \cite{black2019wilderness}.  

The initial build gave an MCD on held out data of 6.483.  While the 2000 best segment dataset gives an MCD of 5.551, which is a large improvement.  The quality of the generated speech goes from understandable, only if you can see the text, to understandable, and transcribable even for non-Mapudungun speakers.

We do not believe we are building the best synthesizer with our current (non-neural) techniques, but we do believe we are selecting the best training data for other statistical and neural training techniques in both speech synthesis and speech recognition.  

\paragraph{Speech Recognition}

For speech recognition (ASR) we used Kaldi~\cite{Povey_ASRU2011}. 
As we do not have access to pronunciation lexica for Mapudungun, we had to approximate them with two settings.
In the first setting, we made the simple assumption that each character corresponds to a pronounced phoneme. In the second setting, we instead used the generated phonetic lexicon also used in the above-mentioned speech synthesis techniques. We had a vocabulary of size 10725 and the test set had 7 out of vocabulary (OOV) words. We approximated the pronunciations for these words using the tools within Kaldi.  
The train/dev/test splits are across conversations, as described above.

Under the first setting, we obtained character error rate of 60\%, and employing the phonemes from generated lexicon, we achieved 30\% phone error rate. For both these systems, we used the nnet training recipe from Kaldi. Naturally, these results are relatively far from the quality of ASR systems trained on large amounts of clean data such as those available in English.
Given the quality of the recordings, and the lack of additional resources, we consider our results fairly reasonable and they would still be usable for simple dialog-like tasks. We anticipate, though, that one could significantly improve ASR quality over our dataset, by using in-domain language models, or by training end-to-end neural recognizers leveraging languages with similar phonetic inventories~\cite{adams2019massively} or by using the available Spanish translations in a multi-source scenario~\cite{anastasopoulos+chiang:interspeech2018}.


\paragraph{Mapudungun--Spanish Machine Translation}
\begin{table}[t]
    \centering
    \begin{tabular}{c|cc|cc}
    \toprule
        Train Data & \multicolumn{2}{c|}{arn$\rightarrow$spa} &  \multicolumn{2}{c}{spa$\rightarrow$arn}\\
        (\#sentences) & BLEU & chrF & BLEU & chrF \\
    \midrule
    \multicolumn{5}{l}{Results on dev set}\\
        220k (all) & 20.98 & 0.5 & 14.12 & 0.4 \\
        100k & 16.89 & 0.4 & 10.93 & 0.3 \\
        50k & 13.70 & 0.4 & 2.05 & 0.1 \\
        10k & 6.26 & 0.3 & 1.09 & 0.1 \\
    \midrule
    \multicolumn{5}{l}{Results on test set}\\
        220k (all) & 20.4 & 0.5 & 12.9 & 0.4 \\
        100k & 16.9 & 0.4 & 6.4 & 0.2 \\
        50k & 13.3 & 0.4 & 1.1 & 0.1 \\
        10k & 7.8 & 0.3 & 0.7 & 0.1 \\
    \bottomrule
    \end{tabular}
    \caption{Machine Translation Results}
    \label{tab:mtresults}
\end{table}

We built neural end-to-end machine translation systems between Mapudungun and Spanish in both directions, using state-of-the-art Transformer architecture \cite{vaswani2017attention} with the toolkit of \cite{nguyen2019transformers}.
We train our systems at the subword level using Byte-Pair Encoding \cite{sennrich2016neural} with a vocabulary of~5000 subwords, shared between the source and target languages.
We use five layers for each of the encoder and the decoder, an embedding size of 512, feed forward transformation size of~2048, and eight attention heads.
We use dropout \cite{srivastava2014dropout} with $0.4$ probability as well as label smoothing set to $0.1$.
We train with the Adam optimizer \cite{kingma2014adam} for up to~200 epochs using learning decay with a patience of six epochs.

The baseline results using different portions of the training set (10k, 50k, 100k, and all (220k) parallel sentences) on both translation directions are presented in Table~\ref{tab:mtresults}, using detokenized BLEU~\cite{papineni2002bleu} (a standard MT metric) and chrF~\cite{popovic2015chrf} (a metric that we consider to be more appropriate for polysynthetic languages, as it does not rely on word n-grams) computed with the sacreBLEU toolkit~\cite{post-2018-call}.
It it worth noting the difference in quality between the two directions, with translation into Spanish reaching~20.4 (almost~21) BLEU points in the development set, while the opposite direction (translating into Mapudungun) shows about a~7 BLEU points worse performance.  This is most likely due to Mapudungun being a polysynthetic language, with its complicated morphology posing a challenge for proper generation.

\section{Conclusion}
With this work we present a resource that will be extremely useful for building language systems in an endangered, under-represented language, Mapudungun. We benchmark NLP systems for speech synthesis, speech recognition, and machine translation, providing strong baseline results.
The size of our resource (142 hours, more than~260k total sentences) has the potential to alleviate many of the issues faced when building language technologies for Mapudungun, in contrast to other indigenous languages of the Americas that unfortunately remain low-resource.

Our resource could also be used for ethnographic and anthropological research into the Mapuche culture, and has the potential to contribute to intercultural bilingual education, preservation activities and further general advancement of the Mapudungun-speaking community.

\section{Acknowledgements}

The data collection described in this paper was supported by
NSF grants IIS-0121631 (AVENUE) and IIS-0534217
(LETRAS), with supplemental funding from NSF's Office of International Science and Education. Preliminary funding for work on Mapudungun was also provided by DARPA
The experimental material is based upon work generously supported by the National Science Foundation under grant 1761548.

\section{Bibliographical References}
\label{main:ref}

\bibliographystyle{lrec}
\bibliography{References}


\end{document}